\documentclass[letterpaper, 10 pt, conference]{ieeeconf}  

\IEEEoverridecommandlockouts                              

\usepackage{cite}
\usepackage{amsmath,amssymb,amsfonts}
\usepackage{algorithmic}
\usepackage{graphicx}
\usepackage{textcomp}
\usepackage{xcolor}
\usepackage{dsfont}
\usepackage{caption} 
\usepackage{subfigure}
\usepackage{url}
\usepackage{booktabs}

\title{\LARGE \bf
PQ-Transformer: Jointly Parsing 3D Objects and Layouts from Point Clouds
}

\author{Xiaoxue Chen$^{1}$, Hao Zhao$^{2}$, Guyue Zhou$^{1}$ and Ya-Qin Zhang$^{1}$
\thanks{$^{1}$ Institute for AI Industry Research (AIR), Tsinghua University, China
        {chenxiaoxue,zhouguyue,zhangyaqin}@air.tsinghua.edu.cn}%
\thanks{$^{2}$Intel Labs China, Peking University, China  
        zhao-hao@pku.edu.cn, hao.zhao@intel.com}%
\thanks{We are grateful to Anker Innovations for supporting this project.}   
}

\begin{document}

\maketitle
\thispagestyle{empty}
\pagestyle{empty}

\begin{abstract}
3D scene understanding from point clouds plays a vital role for various robotic applications. Unfortunately, current state-of-the-art methods use separate neural networks for different tasks like object detection or room layout estimation. Such a scheme has two limitations: 1) Storing and running several networks for different tasks are expensive for typical robotic platforms. 2) The intrinsic structure of separate outputs are ignored and potentially violated. To this end, we propose the first transformer architecture that predicts 3D objects and layouts simultaneously, using point cloud inputs. Unlike existing methods that either estimate layout keypoints or edges, we directly parameterize room layout as a set of quads. As such, the proposed architecture is termed as P(oint)Q(uad)-Transformer. Along with the novel quad representation, we propose a tailored physical constraint loss function that discourages object-layout interference. The quantitative and qualitative evaluations on the public benchmark ScanNet show that the proposed PQ-Transformer succeeds to jointly parse 3D objects and layouts, running at a quasi-real-time (8.91 FPS) rate without efficiency-oriented optimization. Moreover, the new physical constraint loss can improve strong baselines, and the F1-score of the room layout is significantly promoted from 37.9\% to 57.9\%. Code and models can be accessed at \url{https://github.com/OPEN-AIR-SUN/PQ-Transformer}.
\end{abstract}

\section{Introduction}
Recent years have witnessed the emergence of 3D scene understanding technologies, which enables robots to understand the geometric, semantic and cognitive properties of real-world scenes, so as to assist robot decision making. However, 3D scene understanding remains challenging due to the following problems: 1) Holistic understanding requires many sub-problems to be addressed, such as semantic label assignment \cite{b2}, object bounding box localization \cite{b3} and room structure boundary extraction \cite{b1} etc. However, current methods solve these tasks with separate models, which is expensive in terms of storage and computation. 2) The physical commonsense \cite{b5} like gravity \cite{b6} or interference \cite{b7} between different tasks are ignored and potentially violated, producing geometrically implausible results.

\begin{figure*}[h]
  \centering
\subfigure[Input Point Cloud]{
    \label{fig:subfig:subfig-a} 
    \includegraphics[scale=0.23]{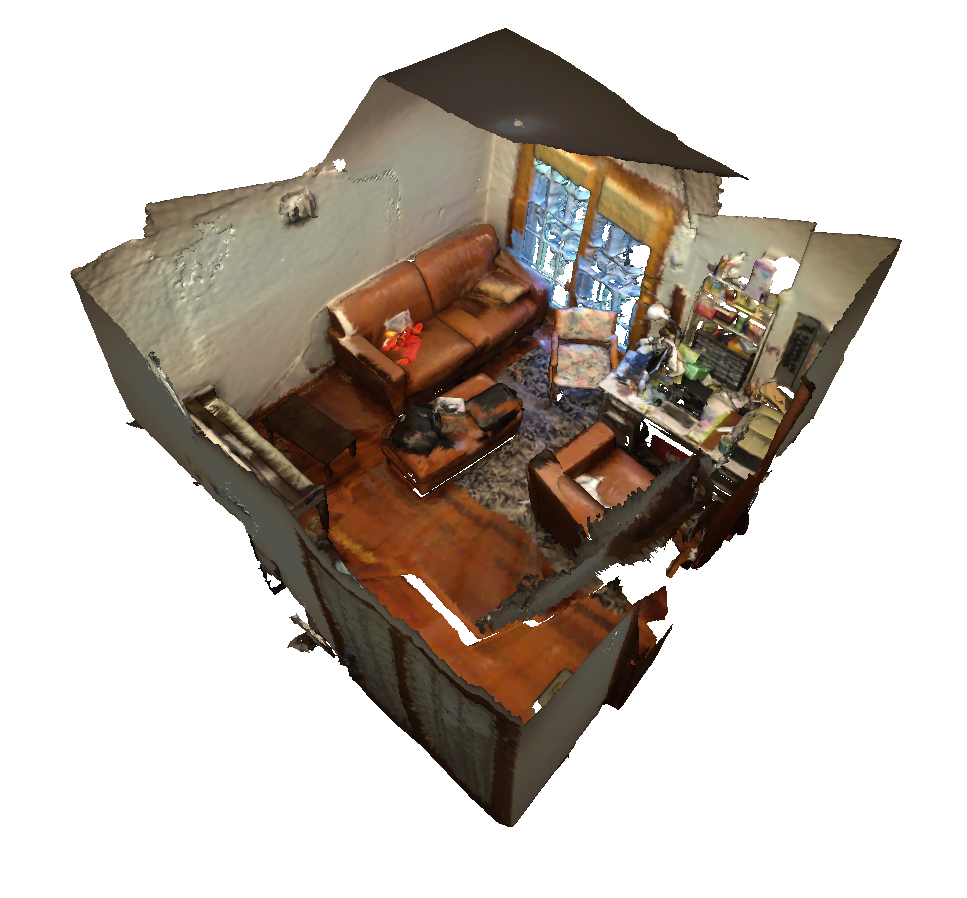}}
  \subfigure[Ground Truth]{
    \label{fig:subfig:subfig-b} 
    \includegraphics[scale=0.23]{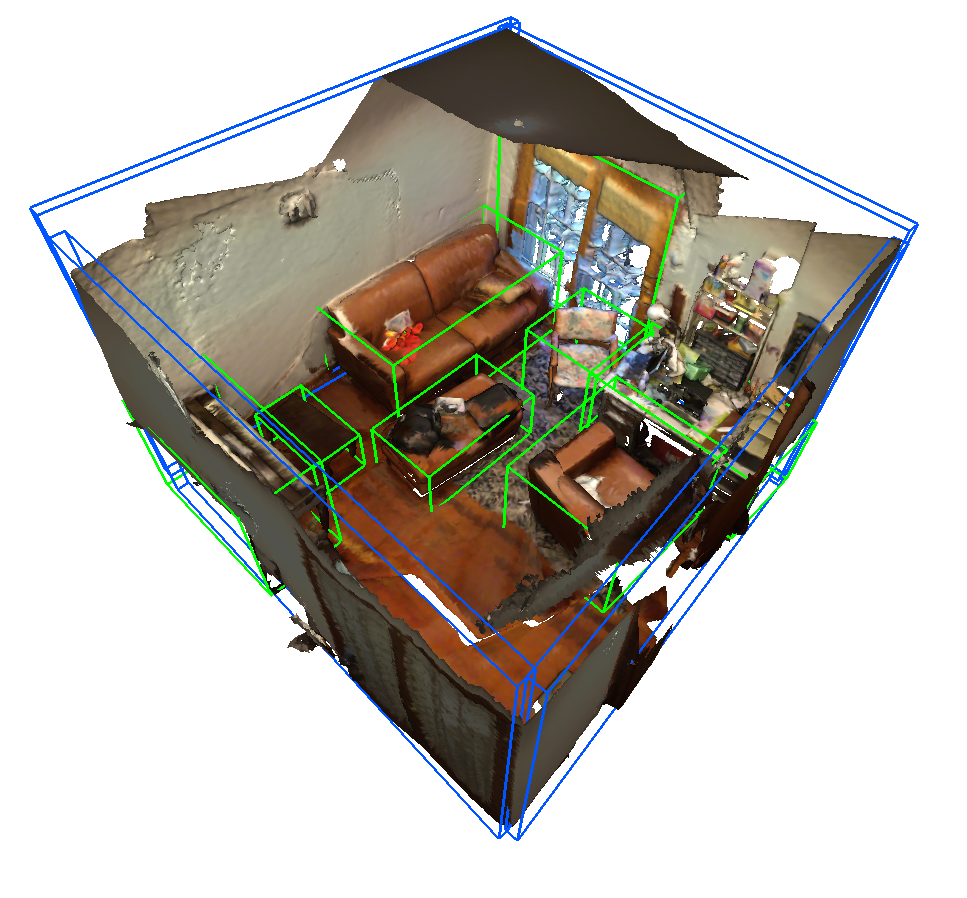}}
  \subfigure[Our Prediction]{
    \label{fig:subfig:subfig-b} 
    \includegraphics[scale=0.23]{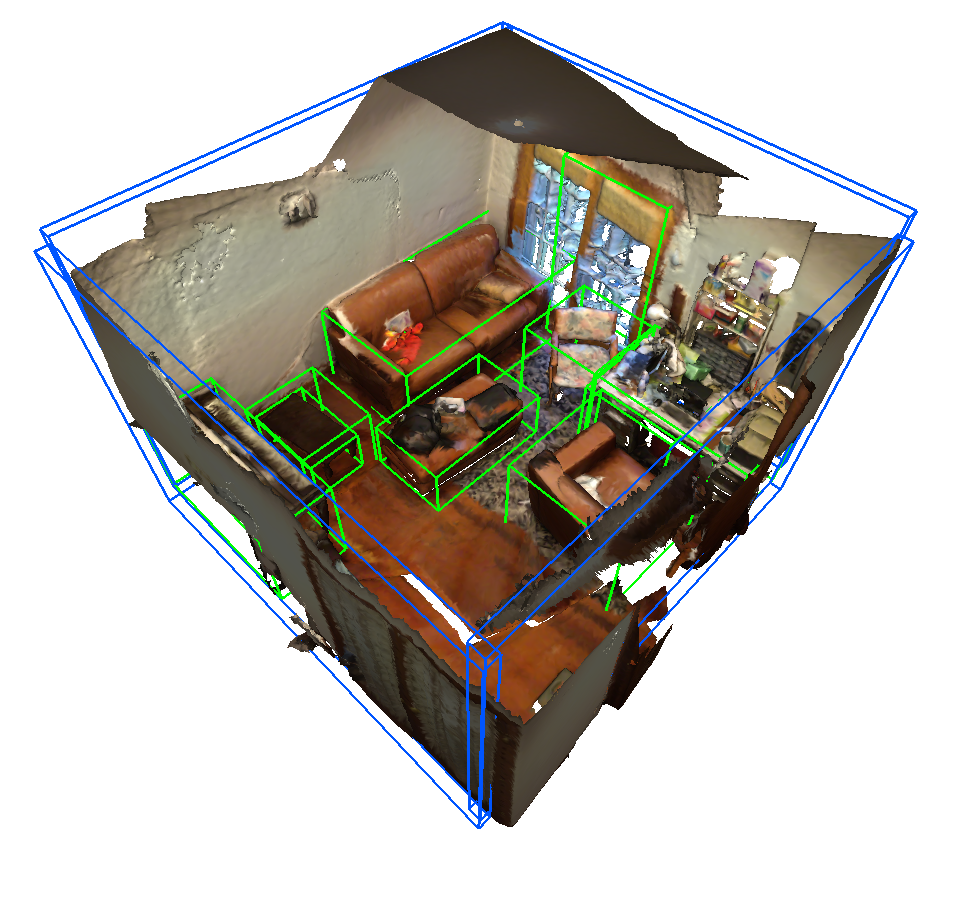}}
  \caption{Illustration of PQ-Transformer on a representative scene of ScanNet. (a) Input point cloud, where the RGB values are not the input, but used for visualization merely. Comparing (b) and (c), the proposed PQ-Transformer succeeds to jointly detect 3D objects (green) and estimates room layouts (blue) in an end-to-end fashion, with high accuracy.}
  \label{fig:teaser} 
\end{figure*}

Aiming for robust 3D scene understanding, we propose PQ-Transformer, the first algorithm that jointly predicts 3D object bounding boxes and 3D room layouts in one forward pass. As illustrated in Fig.~\ref{fig:teaser}(a), the input is 3D point cloud of a scene reconstructed by SDF-based fusion \cite{b8}\cite{b9}. Note that the RGB values are not the inputs, but used for visualization merely. PQ-Transformer predicts a set of 3D object boxes with semantic category labels and another set of quadrilateral (denoted as \emph{quads}) equations representing structure elements (wall, floor and ceiling). Although these quads are of zero width in nature, we set their widths to a small value for better visualization, as illustrated by flat blue boxes in Fig.~\ref{fig:teaser}(c). By comparing to the ground truth in Fig.~\ref{fig:teaser}(b), PQ-Transformer successfully addresses both tasks with high accuracy. Such a joint prediction of 3D objects and layouts is favorable for many robotics applications, since it can largely reduce the overhead for both storage and inference. 

Furthermore, we propose a new loss function by introducing the physical constraints of two tasks during training. This loss function originates from a natural supervision signal,  instead of the human annotated supervisions. Specifically, the interference between object boxes and layout quads are penalized. On one hand, this is consistent with human commonsense and incorporating the constraint makes the learning system closer to the human cognitive system. On the other hand, since trivial mistakes like \emph{objects sinking into the grounds} are corrected, it is natural to expect a more accurate result. Regarding the design of neural network architecture, a transformer is specifically tailored for the joint prediction task. Using two backbones is computationally expensive while using two linear heads leads to contradictory usage on queries. As such, two sets of proposal queries are separately generated for both tasks, striking a balance between efficiency and accuracy.

Benefited from the new representation and network, PQ-Transformer achieves superior performance on challenging scenes of the public benchmark ScanNet. It succeeds to jointly parse 3D objects and layouts, running at a quasi-real-time (8.91 FPS) rate without efficiency-oriented optimization. Moreover, the proposed physical constraint loss improves strong baselines, and the F1-score of the room layout is significantly promoted from 37.9\% to 57.9\%. As demonstrated in Fig.~\ref{fig:qualitative}, the results are useful for both researchers studying 3D scene understanding and practitioners building robotics systems. The technical contributions are summarized as follows.

\begin{itemize}
\item[$\bullet$] PQ-Transformer is the first neural network architecture that jointly predicts 3D objects and layouts from point clouds in an end-to-end fashion.
\item[$\bullet$] Unlike former room layout estimation methods that predict features for layout keypoints, edges or facets, we introduce the quad representation and successfully exploit it for discriminative learning.
\item[$\bullet$] We propose a new physical constraint loss that is tailored for the proposed quad representation, which is principled, generic, efficient and empirically successful.
\item[$\bullet$] PQ-Transformer achieves competitive performance on the ScanNet benchmark. Notably, a new SOTA is achieved for layout F1-score (from 37.9\% to 57.9\%).
\end{itemize}

\section{Related Works}

Many 3D scene understanding tasks are initially defined in the image-based setting, before the advent of commodity RGB-D cameras. \cite{b10} proposes a successful statistical feature called geometric context and parses scenes into geometric categories based upon it, which lays the foundation for pre-deep-learning data-driven 3D scene understanding. Early works \cite{b11}\cite{b12} group line segments according to Manhattan vanishing points and propose primitives for later reasoning, demonstrating the capabilities of estimating room layouts. \cite{b13} shows that proposing 3D objects aligned to the Manhattan frame can be used for joint 3D detection and layout estimation. Later Bayesian models \cite{b14}\cite{b15} are introduced into the field, which model object-layout relationships as statistical priors in a principled manner. \cite{b39} proposes a probabilistic model to jointly parse 3D objects and layouts in urban scenes. In the last decade, many sub-problems benefit from the strong representation power of deep neural networks, including but not limited to object detection \cite{b16}\cite{b17}, object reconstruction \cite{b18}\cite{b19} and room layout estimation \cite{b20}\cite{b21}\cite{b22}\cite{b23}. Recently, joint 3D understanding of several sub-tasks has seen exciting progress, like COPR \cite{b24}, Hoslitic++ \cite{b25} and Total3D \cite{b26}.

After the advent of commodity RGB-D sensors like Kinect or Realsense, 3D scene understanding with point cloud inputs gradually gains popularity. Since the depth information is readily known, scale ambiguity no longer exists. Yet robust understanding is still challenged by issues like occlusion, expensive annotations and sensor noise. SlidingShapes \cite{b27} exploits viewpoint-aware exempler SVMs for 3D detection. DeepSlidingShapes \cite{b28} designs a sophiscated 3D proposal network with data-driven anchors. Semantic Scene completion \cite{b29}\cite{b30} jointly completes scenes and assigns 3D semantic labels, taking a single depth image as the input. Point-wise semantic labelling is successsfully addressed by recently proposed architectures like SparseConv \cite{b31}\cite{b2} or PointNet \cite{b32}. After looking at aforementioned former arts, it is clear that PQ-Transformer is the first transformer-based architecture that jointly predicts 3D objects and layouts from point clouds, with a new quad representation for layouts and its corresponding physical constraint loss.

\begin{figure*}[htbp]
  \centering
  \includegraphics[width=0.83\linewidth]{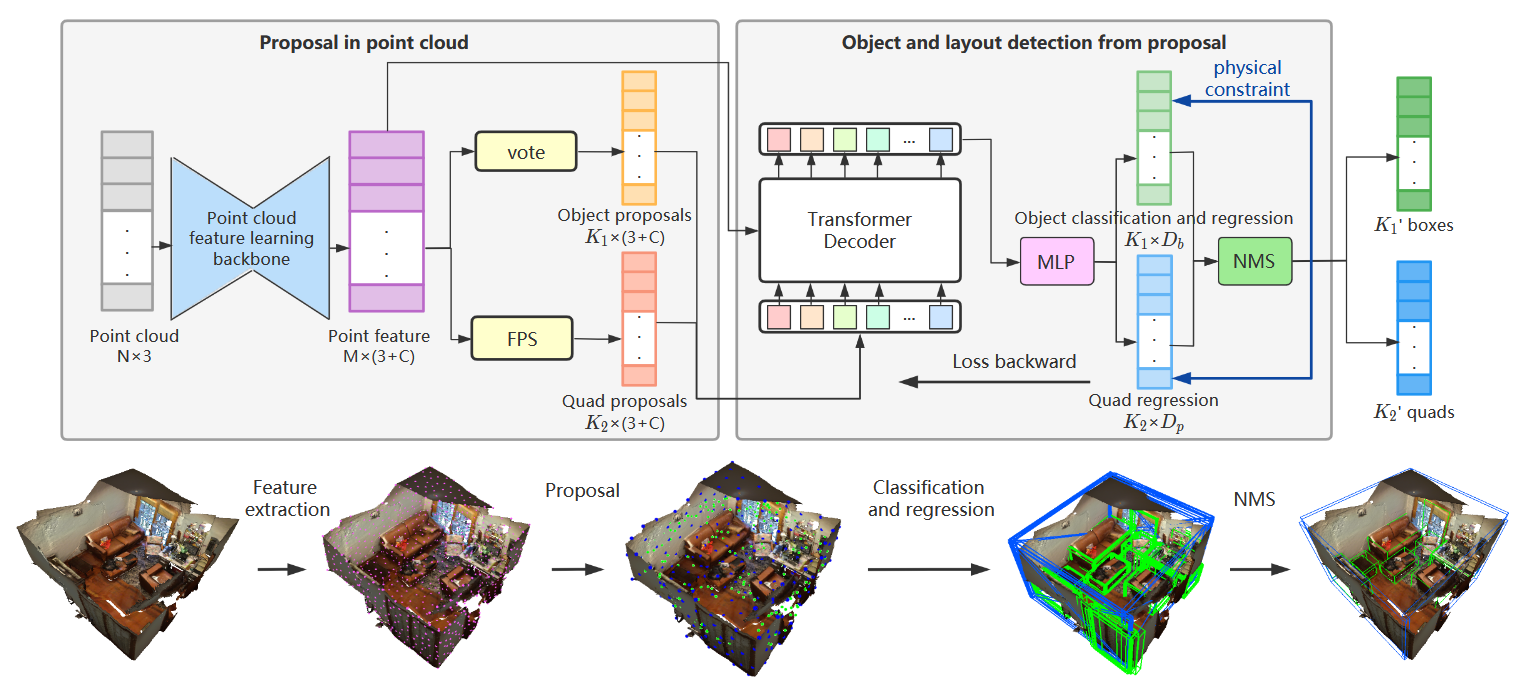}
  \caption{\textbf{Overview of PQ-Transformer}. Given an input 3D point cloud of N points, the point cloud feature learning backbone extracts M context-aware point features of (3+C) dimensions, through sampling and grouping. A voting module and a farthest point sampling (FPS) module are used to generate $K_1$ object proposals and $K_2$ quad proposals respectively. Then the proposals are processed by a transformer decoder to further refine proposal features. Through several feedforward layers and non-maximum suppression (NMS), the proposals become the final object bounding boxes and layout quads.}
  \label{fig:main} 
\end{figure*}

\section{Method}

Our goal is to jointly parse common objects (semantic labels and 3D bounding boxes) and 3D room layouts with a single neural network. To this end, we propose an end-to-end attention-based architecture named PQ-Transformer. We illustrate our architecture in detail with Fig.\ref{fig:main}. 

In the remainder of this section, we first introduce a new representation for 3D room layout, then describe the detailed network architecture. After that, we propose a novel physical constraint loss to refine the joint detection results, by discouraging the interference between objects and layout quads. Finally, we discuss the loss function terms to train PQ-Transformer in an end-to-end fashion.

\subsection{Representation: Layout as Quads}


The representation for 3D object detection is mature and clear. Following former arts \cite{b3}\cite{b33}, we use center coordinate, orientation and size to describe an object bounding box. However, the representation of room layout is still an open problem. Total3D \cite{b26} describes the whole room with a 3D bounding box just like objects. However this representation might not work well because the layout of a real-world room is often non-rectangular. Using a single 3D box isn't enough to accurately describe it. Like image-based layout estimation, SceneCAD \cite{b1} uses layout vertices and edges as the representation. This representation is not compact and requires further fitting to get parametric results.

Different from former methods, we represent the room layouts as a set of quads, which is parametric and compact. Since floors and ceilings are not always rectangular, we only use quads to represent the walls of a room. Then parametric ceiling and floor could be represented by the upper and lower boundaries of the walls. In this way, we formulate the room layout estimation problem into quad detection. Detailed mathematical definition can be found in Section III-C. 


\subsection{Network Architecture}

The overall network architecture is depicted in Fig.\ref{fig:main}. It is composed of four main parts: 1) \textbf{Backbone}: a backbone to extract features from point clouds; 2) \textbf{Proposal modules}: two proposal modules to generate possible objects and layout quads respectively; 3) \textbf{Transformer decoder}: a transformer decoder to process proposal features with context-aware point cloud features; 4) \textbf{Prediction heads}: two prediction heads with several feed forward layers to produce the final predictions, in the joint object-layout output space.


\textbf{Backbone.} We implement the point cloud feature learning backbone with PointNet++ modules. Firstly, four set-abstraction layers are used to down-sample the input 3D point cloud $ P \in \mathbb{R}^{3 \times N}$ and aggregate local point features. Then two feature propagation layers are used to up-sample the points and generate $M$ points with features of dimension $C+3$. Concatenated with coordinates, the extracted features $ f_p \in \mathbb{R}^{M \times \left(C+3\right)} $ are the context-aware local features of the entire scene. It is used as the input to the following proposal modules and the key of cross-attention layers in the transformer decoder.  

\textbf{Proposal modules.} 
We use a voting module and a farthest point sampling (FPS) module to generate proposals for objects and layout quads, respectively. 


Voting. The idea of voting comes from VoteNet\cite{b3}, a technique based on hough voting. Every point in a bounding box is associated with a voting vector towards the box center. To generate votes, we apply a weight-shared multi-layer perceptron (MLP) on $f_p$. The $i$-th point in $f_p$ is represented by feature $s_i=[x_i;f_i]$, with $x_i$ as its 3D point coordinate and $f_i$ as its $C$-dimensional feature. The output of this MLP is offsets of coordinate and feature $[\Delta x_i,\Delta f_i]$. We get its vote $v_i=[y_i,g_i]$, where $y_i=x_i+\Delta x_i$  and $ g_i = f_i+\Delta f_i$. We then sample a subset of $K_1$ votes by using an FPS module on the value of $y_i$. Each cluster is a 3D object proposal.

Farthest Point Sampling. We use FPS to generate $K_2$ initial proposals for layout quads. FPS is based on the idea of repeatedly placing the next sample point in the middle of the least-known area of the sampling domain. FPS starts with a randomly sampled point as the first proposal candidate, and iteratively selects the farthest point from the already selected points until $K_2$ candidates are selected. Though simple, FPS works well for our layout quad detection formulation. Usually the walls are distributed on the outer boundaries of the room and are far from each other. So there is a high probability for FPS to select points on the walls that can provide good enough proposals for quad detection. 



\textbf{Transformer decoder.}  After generating initial proposals based on voting and FPS, we use a transformer decoder to further refine the proposal features. The three basic elements of attention modules are: query ($Q$), key ($K$) and value ($V$), whose dimensions are all $d$ in our case. Proposal features are denoted as $f_c$. First, we feed $f_c$ through self-attention: 
 \begin{align}
& Q=K=V=f_{c} \\
& A  = \emph{\rm softmax}(\frac{Q·K^T}{\sqrt{d}}) \\
& f_{\rm sa}=P_{\rm sa}(AV)+f_{c}
\end{align}

The self-attention layer exploits the mutual relationship between all object and layout quad proposals. 

\begin{figure}[t]
  \centering
  \includegraphics[width=0.7\linewidth]{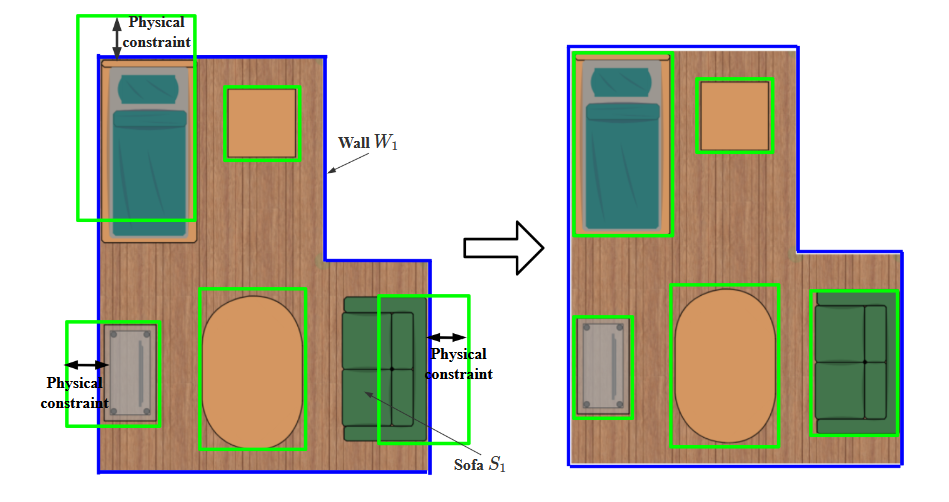}
  \caption{\textbf{The conceptual illustration of physical constraint loss}. The left picture is the detection result without physical constraint loss. As shown in the right picture, after adding physical constraint loss, the bounding box that intersects the wall will move inward. As such the detection result will become more accurate and consistent with physical facts.}
  \label{fig:pc1} 
\end{figure}
 
In addition, we use the context-aware point cloud feature $f_p$ produced by backbone as the key, and fuse it with the proposal features $f_{\rm sa}$ through cross-attention layers:
\begin{align}
& Q = f_{\rm sa},K = V=f_{p} \\
& A = \emph{\rm softmax}(\frac{Q·K^T}{\sqrt{d}}) \\
& f_{\rm ca}=P_{\rm ca}(AV)+f_{\rm sa}
\end{align}


Here $P_{\rm sa}$ and $P_{\rm ca}$ are fully connected layers with batch normalization and ReLU. Our transformer decoder has six blocks, with each one consisted of a self-attention layer and a cross-attention layer. Six blocks generate six sets of detection results, respectively. The detection results of a previous block are used as the position encoding into the current block.

\textbf{Prediction Heads.} After feeding the proposals through the transformer decoder, we use two sets of MLPs as two prediction heads to generate final results. One is used to classify objects and regress object bounding boxes, while the other is used to regress layout quads. For object detection, we follow the formulation of VoteNet\cite{b3}, using a vector of size $2+3 +H+H+S+3S+C$, which consists of $2$ objectness scores, $3$ center regression values, $H$ heading bins, $H$ heading regression values, $S$ size bins, $3S$ size regression values for height-width-depth, and $C$ semantic categories. For layout quad detection, we use a vector of size 10 which is composed of $2$ quadness scores, $3$ center regression values, $2$ size regression values and $3$ normal vector components. Both 3D objects and 2D quads are processed with 3D NMS to remove duplicate boxes, because we give the quad a fixed (but small) width to form a flat cuboid.

\subsection{Physical Constraint}

For now, the object and layout outputs of PQ-Transformer can take unrealistic values, yet there are physical constraints between them in real-world rooms. For example, a table might be near to a wall, but it can never overlap with the wall. In addition, the bottom of the table cannot be lower than the floor. Based on this fact, we design a novel physical constraint loss tailored for our quad representation for layouts, in order to discourage interference. It can help the network generate more precise and reasonable results. It is noteworthy that although there are physical constraints between most objects and walls, some types of objects do overlap with the walls, such as doors, windows and curtains. Therefore, our physical constraints are only designed for those types of objects which will never overlap with walls. We use the set $C_{pc}$ to represent the corresponding object categories. Fig.\ref{fig:pc1} illustrates the role of physical constraints.

\begin{figure}[t]
  \centering
  \includegraphics[width=0.6\linewidth]{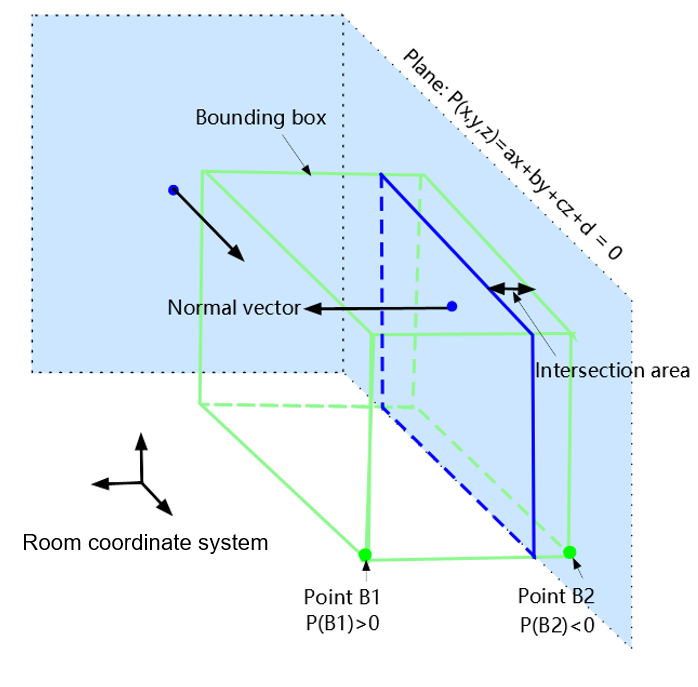}
  \caption{\textbf{The mathematical illustration of physical constraints}. We use the vertices of the object bounding box (green) to justify whether it intersects with the wall (blue). The equation of wall divides the space into two parts. B2 is out of the room and B1 is inside the room.}
  \label{fig:pc} 
\end{figure}

\begin{figure*}[htbp]
  \centering
  \includegraphics[width=0.8\linewidth]{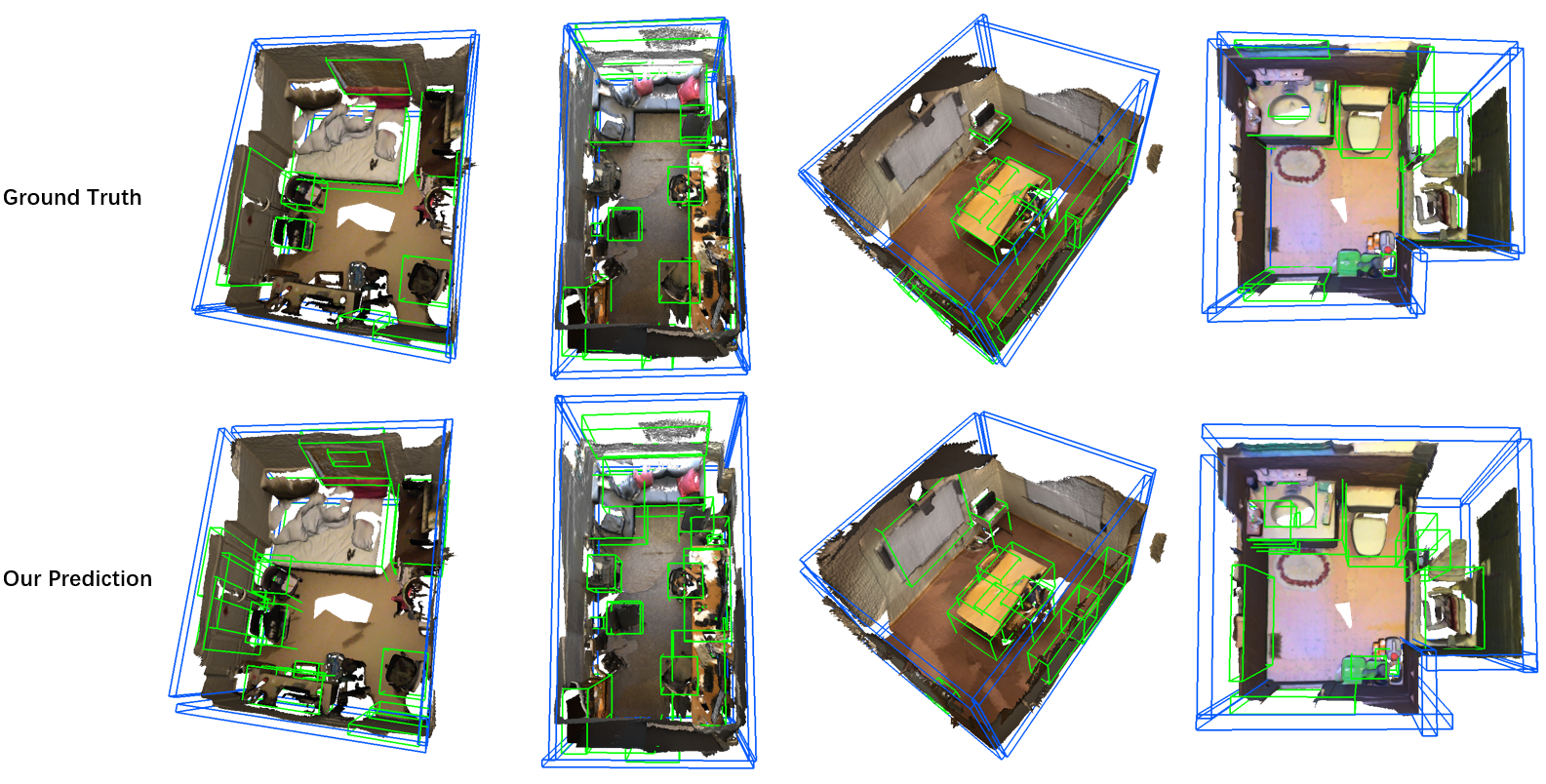}
  \caption{\textbf{Qualitative prediction results on ScanNet}. Objects are outlined in green while layout quads are outlined in blue.}
  \label{fig:qualitative} 
\end{figure*}

We use a quad in the 3D Euclidean space to represent a wall, and the quad defines a 3D plane whose equation is:
\begin{align}
ax+by+cz+d=0
\end{align}

Vector $(a,b,c)$ is exactly the normal vector of the plane. For dis-ambiguity, we make all normal vectors point to the center of the room manually. We could divide the 3D space into two parts using this plane. For a point with coordinate $(x',y',z')$, if $ax'+by'+cz'+d>0$, the point is at the same side of the room center, otherwise, the point is out of the room. Fig.\ref{fig:pc} illustrates the situation that a bounding box (green) intersects with a wall (blue, right) and vertice B1 is in the room while vertice B2 is out of the room. For a 3D object box, we traverse its eight vertices, determine whether they intersect the walls using the plane equations. 

For a vertice $(x_i,y_i,z_i)$, the physical constraint loss we minimize takes the form of ReLU$\left[ -(ax_i+by_i+cz_i+d) \right]$. However, imposing this loss on all objects and walls might cause wrong constraints. For example, in the left part of Fig.\ref{fig:pc1}, wall $W_1$ and sofa $S_1$ should not constrain each other since the bounding box of $S_1$ and the quad of $W_1$ actually do not intersect. But if we impose the loss equation above between $W_1$ and $S_1$, it leads to a no-zero physical constraint loss. To avoid this kind of wrong constraints, we first determine whether the projection of a bounding box vertice is within the wall quad before calculating the physical constraint loss. We project the vertice onto the wall plane, and compare its projection with the quad size. The loss equation for a set of detection results with $K_1$ objects and $K_2$ quads is:
\begin{align}
  \mathcal{L}_{pc}  = & \sum_{\substack{i=1\\C_i \in C_{pc}}}^{K_1} \sum_{j=1}^{K_2} \sum_{p=1}^8 \emph{\rm ReLU}\left[ -(a_jx_{ip}+b_jy_{ip}+c_jz_{ip}+d_j) \right] \notag\\
  &\mathds{1}[\Pi_{q_j}(x_{ip},y_{ip},z_{ip})\; \rm in\; q_j]
\end{align}

$q$ denotes a quad. $\Pi_{q_j}$ means the operator projecting a point onto the plane that $q_j$ defines. $C_i$ means the sementic class of object $i$ and $C_{pc}$ is the set of object classes to calculate physical constraint loss. The plane equation of $j$-th quad is $a_jx+b_jy+c_jz+d=0$ and $\mathds{1}[\Pi_{q_j}(x_{ip},y_{ip},z_{ip})\; \rm in\; q_j]$ indicates whether the projection of vertice $(x_{ip},y_{ip},z_{ip})$ is in $q_j$: if it is, return 1; otherwise, return 0.


\subsection{Loss}

First we denote the layer number of transformer decoder as $L$. We get $L+1$ sets of detection results in total. Specifically, $L$ sets are generated from $L$ layers of the decoder and one set is generated from the proposal module. Then we calculate loss on each set of results and use the summation as the final loss. Losses on intermediate decoder outputs and proposal module outputs play the role of auxiliary supervision, which help PQ-Transformer converge. Let the loss of the $i$-th set of detection results be $\mathcal{L}_{i}$, the total loss used in training is: 
\begin{align}
	\mathcal{L}_{\rm total}=\mathcal{L}_{\rm pc} + \mathcal{L}_{\rm vote} + \frac{1}{L+1} \sum_{i=1}^{L+1} \mathcal{L}_{i}
\end{align}

Here $\mathcal{L}_{\rm vote}$ is the loss for voting vectors:

\begin{align}
	\mathcal{L}_{\rm vote}=\frac{1}{M} \sum_{i=1}^M \| \Delta x_i-\Delta x_i^* \| \mathds{1}[\rm s_i\; on\; object]
\end{align}
Here $\Delta x_i^*$ is the ground truth voting vector. $\mathds{1}[\rm s_i\; on\; object]$ indicates whether the point $s_i$ is inside a bounding box. If it is, the value is 1, otherwise it is 0. 
For each set of results:
\begin{align}
	\mathcal{L}_{i}= \mathcal{L}_{\rm object} + \mathcal{L}_{\rm quad}
	\label{con:pc_loss} 
\end{align}
$\mathcal{L}_{\rm object}$ is the loss between predicted bounding boxes and ground truth boxes, while $\mathcal{L}_{\rm quad}$ is the loss between predicted quads and ground truth quads. They are calculated as below: 
\begin{align}
	\mathcal{L}_{\rm object}=\lambda_1\mathcal{L}_{\rm objectness} + \lambda_2\mathcal{L}_{\rm box} + \lambda_3\mathcal{L}_{\rm cls}
\end{align}
\begin{align}
	\mathcal{L}_{\rm quad} & =\lambda_4\mathcal{L}_{\rm quadness} + \lambda_5\mathcal{L}_{\rm quad\_center} \notag\\
	& + \lambda_6\mathcal{L}_{\rm quad\_normal\_vector} + \lambda_7\mathcal{L}_{\rm quad\_size}
\end{align}
$\lambda_1 \sim \lambda_7$ are loss weight parameters. We use cross entropy loss for all classification results like $\mathcal{L}_{\rm objectness}$ and $\mathcal{L}_{\rm cls}$. For regression results like $\mathcal{L}_{\rm quad\_center}$ and $\mathcal{L}_{\rm quad\_size}$, we use smooth L1 loss. Detailed loss weight settings can be found in the supplementary material.

\section{Experiment}


\subsection{Comparisons with State-of-the-art Methods}

\textbf{Evaluation Details.} We validate PQ-Transformer on the widely-used indoor scene dataset ScanNet \cite{b34}. It contains $\sim$1.2K real-world RGB-D scans collected from hundreds of different rooms. It is annotated with semantic and instance segmentation labels for 18 object categories. In addition, SceneCAD\cite{b1} introduces a new dataset by adding 3D layout annotations to ScanNet, allowing large-scale data-driven training for layout estimation. The SceneCAD layout dataset contains $\sim$13.8k corners, $\sim$20.5K edges and $\sim$8.4K polygons. We first preprocess these annotations, choosing polygons which have 4 vertices and nearly horizontal normal vectors as the ground truth of wall quads during training. We use the official ScanNet data split. In later paragraphs and tables, \emph{single} means that we train object detector and layout estimator separately, and \emph{joint} represents our full PQ-Transformer architecture illustrated in Fig.~\ref{fig:main}.

\begin{figure}[t]
  \centering
  \includegraphics[width=0.9\linewidth]{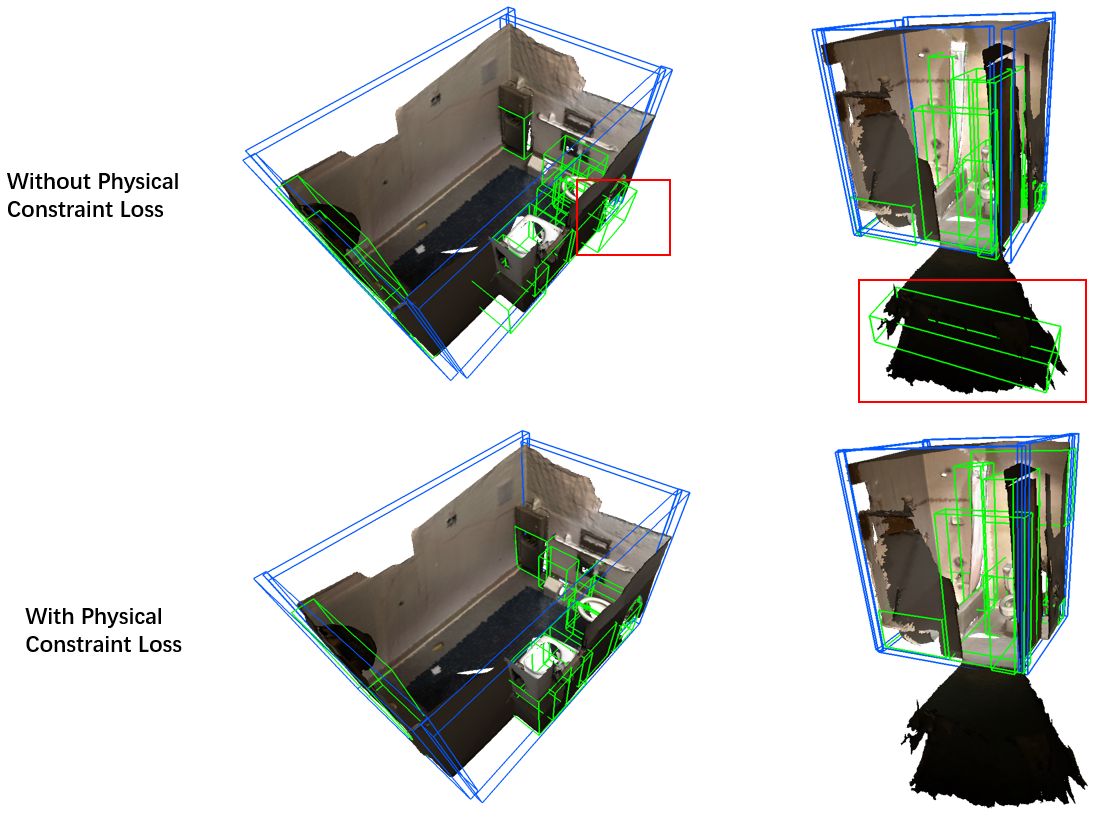}
  \caption{\textbf{Qualitative comparisons on ScanNet}. After adding the physical constraint loss, the bounding boxes no longer overlap with the walls (left), and the wrongly predicted bounding box disappears (right).}
  \label{fig:comparison} 
\end{figure}

\textbf{Layout estimation.} We show our layout estimation results on ScanNet in Tab.~\ref{tab:layoutdetection}. SceneCAD\cite{b1} uses a bottom-up pipeline to predict quads hierarchically. Contrasting with SceneCAD, our approach generates quad proposals directly and refines them with transformer. For comparison, we use the same evaluation metrics as SceneCAD does. As mentioned before, ceiling and floor polygons (not necessarily quads) are generated by connecting the upper and lower boundaries of predicted wall quads (see details in the supplementary material). Polygon corners are considered successfully detected if the predicted corner is within a radius of 40 $cm$ from any ground truth corner. Similarly, predicted polygons are considered correct if composed by the same corner set as any ground truth polygon. As shown in the Tab.\ref{tab:layoutdetection}, the room layout F1-score on ScanNet is significantly promoted from 37.9\% to 57.9\%. And if only considering wall quads, the F1-score is 70.9\%. For joint detection, the F1-score also outperforms previous state-of-the-art by 17.9\%. 

\begin{table}[htbp]
\caption{Layout estimation results on ScanNet.}
\begin{center}
\begin{tabular}{ccc}
\toprule
Method & F1-score (all) & F1-score (wall only)  \\
\midrule
SceneCAD & 37.9$^1$ &   $\backslash$ \\
\midrule
Ours (joint) & 55.8 & 68.7 \\
Ours (single) & \textbf{57.9} & \textbf{70.9} \\
\bottomrule
\end{tabular}
\label{tab:layoutdetection}
\end{center}
\end{table}

\footnotetext[1]{This result is taken from: https://www.ecva.net/papers/eccv\_2020/papers\_E\\CCV/papers/123670596.pdf (page 11, line 1).} 

\textbf{3D object detection.} We compare our 3D object detection results with previous state-of-the-arts in Tab.\ref{objectdetection}. L6 means 6 attention layers and O256 means 256 proposals. HGNet \cite{b35} exploits a graph convolution network based upon hierarchical modelling, for 3D detection. VoteNet \cite{b3} uses point-wise voting vectors to generate object proposals. Group-Free \cite{b33} is an attention-based detector that generates object proposals with k-nearest point sampling. GSDN \cite{b38} uses a fully convolutional sparse-dense hybrid network to generate the support for object proposals. H3DNet \cite{b37} predicts a diverse set of geometric primitives and converts them into object proposals. Following the standard evaluation protocol, we use mean Average Precision (mAP) to evaluate PQ-Transformer on object detection. Tab.\ref{objectdetection} shows that our approach performs comparable with the state-of-the-art methods.

\begin{table}[htbp]
\caption{3D object detection results on ScanNet.}
\begin{center}
\begin{tabular}{cc}
\toprule
Method &  mAP@0.25 \\
\midrule
VoteNet \cite{b3} & 58.7  \\
HGNet \cite{b35} & 61.3   \\
GSDN \cite{b38} & 62.8   \\
H3DNet \cite{b37} & 67.2  \\
Group-Free \cite{b33} (L6, O256) & \textbf{67.3}   \\
\midrule 
Ours (joint, L6, O256) & 66.9   \\
Ours (single, L6, O256) & 67.2  \\
\bottomrule
\end{tabular}
\label{objectdetection}
\end{center}
\end{table}

\subsection{Ablation Study}

\textbf{Physical constraint loss.} To investigate the necessity of physical constraint loss, we train two models with and without it. We demonstrate the results in Tab.\ref{pc_result}. The mAP of object detection rises from 64.4\% to 66.9\% after adding physical constraint loss and the F1-score of layout estimation increases from 54.7\% to 55.8\%, which clearly shows the effectiveness of our physical constraint loss. We also show the number of collisions between objects and walls with two models in Tab.\ref{pc_result}. One collision means a vertex of the object bounding box is out of the room. The sharp drop in the number of collisions shows that our physical constraint loss discourages object-layout interference successfully. 

As demonstrated in Fig.\ref{fig:comparison}, the object detection results are more reasonable with physical constraint loss. In the top-left sample, the bounding box of the toilet in the red box intersects with the wall, which is impossible in the real-world. While training with the physical constraint loss, this error no longer exists. In the top-right sample, influenced by the point cloud outside the room, there is a meaningless bounding box there when training without physical constraint loss. And it vanishes after adding the loss.

\begin{table}[htbp]
\caption{Joint prediction results on ScanNet with or without using the physical constraint loss.}
\begin{center}
\begin{tabular}{cccc}
\toprule
 & Object (mAP) & Layout (F1-score) & No. Collisions\\
\midrule
w/o $\mathcal{L}_{pc}$ & 64.4 & 54.7 & 7208 \\
w/ $\mathcal{L}_{pc}$  & \textbf{66.9} & \textbf{55.8} &\textbf{9} \\
\bottomrule 
\end{tabular}
\label{pc_result}
\end{center}
\end{table}

 \textbf{Auxiliary loss.} In Tab.~\ref{loss}, we validate the effectiveness of the auxiliary losses on intermediate decoder outputs (denoted as \emph{intermediate})  and proposal module outputs (denoted as \emph{proposal}). Removing \emph{intermediate} loss reduces the F1-score of layout estimation by 8.0\%. And removing \emph{proposal} loss or both losses also lead to large performance drops. This fact demonstrates that these two losses serve as critical auxiliary supervision roles during the training process. Specifically, \emph{proposal} losses guide the voting modules and FPS modules to generate refined proposals and \emph{intermediate} losses propagate gradients to early transformer layers, which help PQ-Transformer converge towards superior performance.
  
\begin{table}[htbp]
\caption{Joint prediction results on ScanNet with different losses.}
\begin{center}
\begin{tabular}{ccc|cc}
\toprule
 Proposal & Intermediate & Last layer &Object &Layout\\
\midrule
  & & \checkmark & 44.0 & 44.9  \\
 \checkmark & & \checkmark & 46.7 & 47.8  \\
  & \checkmark &  \checkmark& 64.4 & 52.9  \\
 \checkmark &  \checkmark&  \checkmark& \textbf{66.9} & \textbf{55.8} \\
\bottomrule 
\end{tabular}
\label{loss}
\end{center}
\end{table}

\textbf{Architecture.} Is transformer necessary in our model? We conduct experiments to understand its effect. For comparison, we train a naive joint detection model without the transformer decoder, by feeding the proposals into the two prediction heads directly after the voting module and FPS module. As shown in Tab. \ref{transformer}, PQ-transformer outperforms the model without transformer by 4.2\% in object detection and 20.4\% in layout estimation. We believe the reason is that transformer can leverage the spatial relations between all proposals, and naturally refine them with context-aware point features through cross-attention, especially for layout proposals that are strongly related to each other. 
\begin{table}[htbp]
\caption{results with or without transformer decoder.}
\begin{center}
\begin{tabular}{c|cc}
\toprule
Architecture & Object mAP & Layout F1-score \\
\midrule
w/o transformer & 62.7  & 35.4  \\
PQ-Transformer & \textbf{66.9}  & \textbf{55.8}  \\
\bottomrule
\end{tabular}
\label{transformer}
\end{center}
\end{table}

Since how to design a single transformer for two structured prediction tasks remains unclear, we design experiments to compare several alternative architectures which are shown in Tab.\ref{architecture}. \emph{one proposal} represents the model trained with a single proposal module for both object detection and layout estimation. In this case, our two tasks would compete for bottom-up proposals. And our architecture depicted in Fig.\ref{fig:main} is denoted as \emph{two proposals}. We present the size of different models. Tab.\ref{architecture} shows that although \emph{single} has achieved the best results, its runtime speed is very slow and it has the largest model size. \emph{one proposal} has the best efficiency in terms of both speed and storage, but its performance is obviously poor. Our model has achieved comparable quantitative results with \emph{single} while the speed and storage are close to \emph{one proposal}. This verifies the effectiveness of our architecture and shows its superiority in storage and computation compared with separately trained models. We believe this insight is useful for similar multi-task transformer architectures: separating different tasks at the proposal stage, rather than inputs or prediction heads.

\begin{table}[htbp]
\caption{Architecture design comparisons.}
\begin{center}
\begin{tabular}{c|c|c|cc}
\toprule
 & Speed & Size  & Object & Layout \\
Architecture & (FPS) & (MB) &mAP & F1-score \\ 
\midrule
single &4.29  & 122.7 &\textbf{67.2} & \textbf{57.9}  \\
one proposal & \textbf{9.52}  & \textbf{65.7} & 44.6 & 52.4  \\
two proposals & 8.91  & 67.6 & 66.9 & 55.8  \\
\bottomrule
\end{tabular}
\label{architecture}
\end{center}
\end{table}

\subsection{Qualitative Results and Discussion}


\begin{figure}[t]
  \centering
  \includegraphics[width=0.8\linewidth]{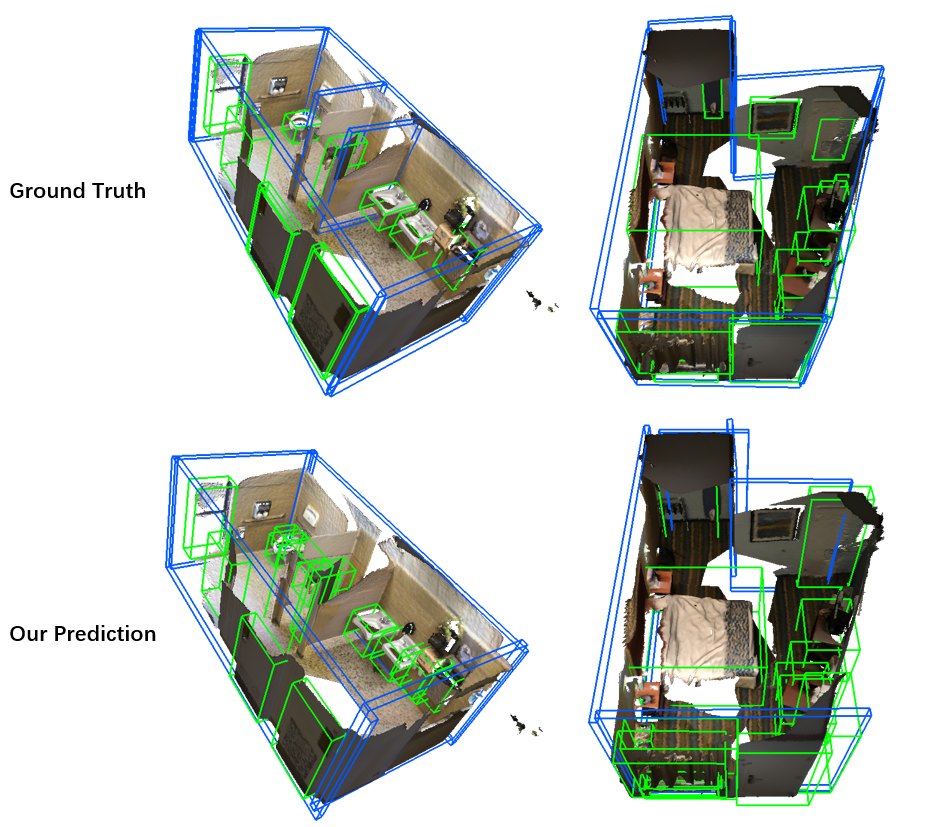}
  \caption{\textbf{Failure cases on ScanNet}. PQ-Transformer fails to detect the two partition walls in the middle of the room (left) and the inclined wall (right).}
  \label{fig:failure} 
\end{figure}


Fig.\ref{fig:qualitative} shows our joint parsing results on ScanNet. It is manifest from Fig.\ref{fig:qualitative} that our approach can predict the wall quads precisely even if the room is non-rectangular and detect the bounding boxes of most objects successfully. The differences between our object detection results and ground truth mainly arise from annotation ambiguity and duplicate detection. To be more exact, in the first column of Fig.\ref{fig:qualitative}, our approach detects the desk in the bottom-left corner while it isn't annotated in the ground truth. And in the second column, our approach recognizes the corner sofa as two separate sofas while ground truth takes it as a whole one. More qualitative results are provided in the supplementary material. Considering the diversity of these scenes, we believe PQ-Transformer is accurate enough for various robotics applications.

Our layout estimation approach still has limitations. Fig.\ref{fig:failure} shows some failure cases on ScanNet. In the first column, our approach fails to detect the two partition walls in the middle of the room. And we are unable to detect the inclined wall on the right side of the room, in the second column. We observe that PQ-transformer performs better on the outer walls and walls aligned with canonical directions. This is because that the number of partition walls and inclined walls is small in the dataset, and the unbalanced distribution of the training data makes our model biased towards canonical wall settings. This may be solved with a more balanced layout dataset. In addition, inclined walls and partition walls may overlap with the object bounding boxes, violating the formulation of the proposed physical constraint loss. 


\section{Conclusion}
 In this study, we develop the first attention-based neural network to predict 3D objects and layout quads simultaneously, taking only point clouds as inputs. We introduce a novel representation for layout: a set of 3D quads. Along with it, we propose a tailored physical constraint loss function that discourages object-layout interference. A multi-task transformer architecture that strikes the balance between accuracy and efficiency is proposed. We evaluate PQ-Transformer on the public benchmark ScanNet and show that: 1) The new physical constraint loss can improve strong baselines. 2) The layout F1-score on ScanNet is significantly boosted from 37.9\% to 57.9\%. We believe our method is useful for robotics applications as the final model runs at a quasi-real-time (8.91 FPS) rate without efficiency-oriented optimization.



\newpage

\section*{APPENDIX}

\subsection{Method Details}

This section provides additional implementation details of PQ-Transformer. First, we show network architecture in section A-1 and layout estimation details in section A-2. Then in section A-3, we discuss our implementation of physical constraint loss. After that, we elaborate on our loss weights setting in section A-4. And finally, we provide training details in section A-5.

\subsubsection{Architecture Specification Details}
The point cloud feature learning backbone is implemented with modules in PointNet++\cite{b2}. It consists of 4 set abstraction layers and two feature propagation layers. For each set abstraction layer, the input point cloud size is down-sampled to 2048, 1024, 512 and 256 respectively. And the two feature propagation layers up-sample the point cloud features to 512 and 1024 by applying trilinear interpolation on the input features.

We generate $K_1$ proposals for object detection and $K_2$ for layout estimation where $K_1 = K_2 = 256$. Then we use a transformer decoder with 6 attention layers to refine proposals. The head number of it is 8.

Following \cite{b1}, we parameterize an oriented 3D bounding box as a vector of size  $2+3+H+H+S+3S+C$. The first two are objectness scores and the next three are center regression results. H is the number of heading bins. We predict a classification score and a regression offset for each heading bin. S is the number of size bins. Similarly, we predict a classification result and three regression results (height, width and length) for each size bin. And C is the number of semantic classes. In ScanNet\cite{b5}, we set H = 12 and S = C = 18. 

\subsubsection{Layout Estimation Details: Quad NMS and Ceiling/Floor Evaluation}

The layout estimation result obtained by the prediction head contains $K_2$ quads. To remove duplicate quads, we  give each quad a fixed width to form a flat cuboid. In our implementation, we set the width to 10 $cm$ so that we could process these flat cuboids easily with 3D NMS. We set the IoU threshold of NMS to 0.25, during training.

\begin{figure*}[htbp]
  \centering
  \includegraphics[width=0.95\linewidth]{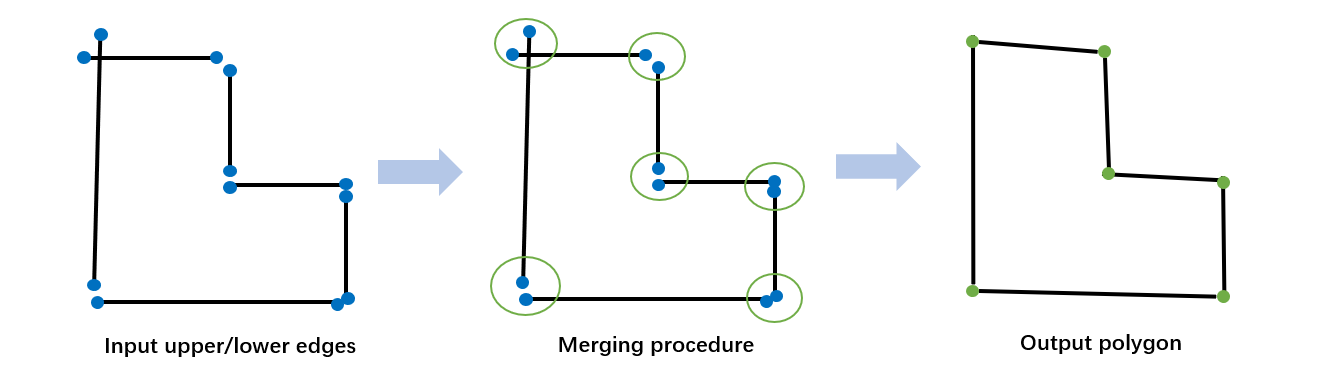}
  \caption{Illustration of the merging procedure in Ceiling/Floor evaluation.}
  \label{fig:merging} 
\end{figure*}

Through 3D NMS and quadness filtering (only consider quads with quadness scores \textgreater 0.5), we get $K_2'$ quads. We use them to generate ceiling and floor. First, We initialize a list $E_c$ for the ceiling. Then we traverse $K_2'$ quads, adding the upper edge of the quad into $E_c$. After that we iterate through $E_c$. If the distance between vertices on two edges in $E_c$ is less than 40 $cm$, the two vertices will be merged by averaging. After merging, $E_c$ becomes the edge estimation of the ceiling. which defines a polygon. The merging procedure is illustrated in Fig \ref{fig:merging}. Similarly, we calculate floor edges using the lower edges of quads in the same way. As mentioned in the main paper, we use the same evaluation metrics as SceneCAD\cite{b6} does. Two vertices are considered the same if they are within 40 $cm$ of each other. And an edge is considered correct if composed by the same two vertices with any ground truth edge. So if the ceiling or floor is composed of the same edges with any ground truth polygon, it is considered successfully estimated. 

\subsubsection{Physical Constraint Implementation Details}

The computational complexity of the generic physical constraint loss we introduced in the main paper is high. Because we have to traverse eight vertices of all object bounding boxes and all quads. Considering the fact that the wall quads are nearly vertical, we only calculate a 2D version of physical constraint loss in practice to reduce computation. We transform all the bounding boxes and quads into top-down view. Then the bounding boxes become rectangles and the quads become line segments. We represent a line segment with equation $ax+by+d=0$ and a length $l$. And for a vertice $(x_i,y_i)$, the physical constraint loss between it and the line segment becomes ReLU$\left[ -(ax_i+by_i+d) \right]$. To further improve efficiency, we accelerate iteration of vertices with matrix operations. We use $P \in \mathbb{R}^{n \times 2}$ to describe $n$ vertices whose coordinates are of dimension $2$. Q = $[a,b]^T$ represents the normal vector of the line segment. $\Pi \in \mathbb{R}^{n}$ indicates whether the projection of vertices are in the line segment. The loss between n vertices and the line segment (quad) is: 
\begin{align}
  \mathcal{L}_{pc}  = \emph{\rm sum} \left( \emph{\rm ReLU}\left[ -(PQ+d) \right] \odot \Pi \right)
\end{align}

where $\odot$ denotes element-wise product and \emph{sum} means summation of all elements in the matrix. We compare the training time of different $\mathcal{L}_{pc}$ implementations of in Tab. \ref{timecost}, which shows the efficiency of the our implementation.

\begin{table}[htbp]
\caption{Time cost for one batch (8 samples) training.}
\begin{center}
\begin{tabular}{cc}
\hline
 & \multicolumn{1}{|c}{Time cost /s} \\
\cline{1-2}
w/o $\mathcal{L}_{pc}$ & \multicolumn{1}{|c}{\textbf{0.24}}  \\
w/ $\mathcal{L}_{pc}$ (trivial) & \multicolumn{1}{|c}{1.77}  \\
w/ $\mathcal{L}_{pc}$ (efficient) & \multicolumn{1}{|c}{0.32}  \\
\hline
\end{tabular}
\label{timecost}
\end{center}
\end{table}

\subsubsection{Loss Balancing Details}
PQ-Transformer is trained with a multi-task loss in an end-to-end fashion. As mentioned in the main paper, the object loss $\mathcal{L}_{\rm object}$ is denoted as:
\begin{align}
	\mathcal{L}_{\rm object}=\lambda_1\mathcal{L}_{\rm objectness} + \lambda_2\mathcal{L}_{\rm box} + \lambda_3\mathcal{L}_{\rm cls}
\end{align}
We use the loss weights to balance different loss functions as follows:
$$\lambda_1 = 0.5, \quad \lambda_2 = 1, \quad \lambda_3 = 0.1.$$
The quad loss $\mathcal{L}_{\rm quad}$ is denoted as:
\begin{align}
	\mathcal{L}_{\rm quad} = & \lambda_4\mathcal{L}_{\rm quadness} + \lambda_5\mathcal{L}_{\rm quad\_center} \notag \\
	& + \lambda_6\mathcal{L}_{\rm quad\_normal\_vector} + \lambda_7\mathcal{L}_{\rm quad\_size}
\end{align}	
And the weights for $\mathcal{L}_{\rm quad}$ are:
$$\lambda_4 = 0.5,\quad \lambda_5 = \lambda_6 = \lambda_7 = 1.$$
Specifically, the detailed form of $\mathcal{L}_{\rm box}$  is:
\begin{align}
	\mathcal{L}_{\rm box} = & \lambda_8  \mathcal{L}_{\rm center} + \lambda_9  \mathcal{L}_{\rm head\_cls} +
	 \lambda_{10} \mathcal{L}_{\rm heading\_reg} \notag\\
	&+ \lambda_{11}  \mathcal{L}_{\rm size\_cls}+ \lambda_{12} \mathcal{L}_{\rm size\_reg} 
\end{align}	
where $\mathcal{L}_{\rm center}$ is loss for bounding box center, $\mathcal{L}_{\rm head\_cls}$ is heading bin classification loss, $\mathcal{L}_{\rm heading\_reg}$ is heading bin regression loss, and $\mathcal{L}_{\rm size\_cls}$ and $\mathcal{L}_{\rm size\_reg}$ are classification score loss and regression loss for box size bin respectively. And the weights are as follows:

$$ \lambda_8 = \lambda_{10} = \lambda_{12} = 1,\quad \lambda_9 = \lambda_{11} = 0.1.$$

\subsubsection{Training Details}
We train PQ-Transformer with with three NVIDIA GeForce RTX 3090 GPUs and test it on a single GPU. The network is trained with an AdamW optimizer in an end-to-end fashion. And we sample 40K vertices from ScanNet as our input point clouds, setting batch size per GPU to 8. We spend 600 epochs to train the model.

\subsection{More Qualitative Results}
We show more qualitative results of PQ-Transformer on ScanNet. The results are shown in Fig.\ref{fig:more}. Considering the diversity of scenes and objects in these cases, we believe our approach has achieved accurate and robust object detection and layout estimation.

\subsection{ScanNet Per-category Evaluation}

Tab. \ref{objectdetection} demonstrates per-category average precision on ScanNet with a 0.25 IoU threshold. It shows that PQ-Transformer performs comparable with state-of-the-art methods and performs better in some categories.

\begin{figure*}[h]
  \centering
    \subfigure{
    \label{fig:subfig:subfig-a} 
    \includegraphics[scale=0.5]{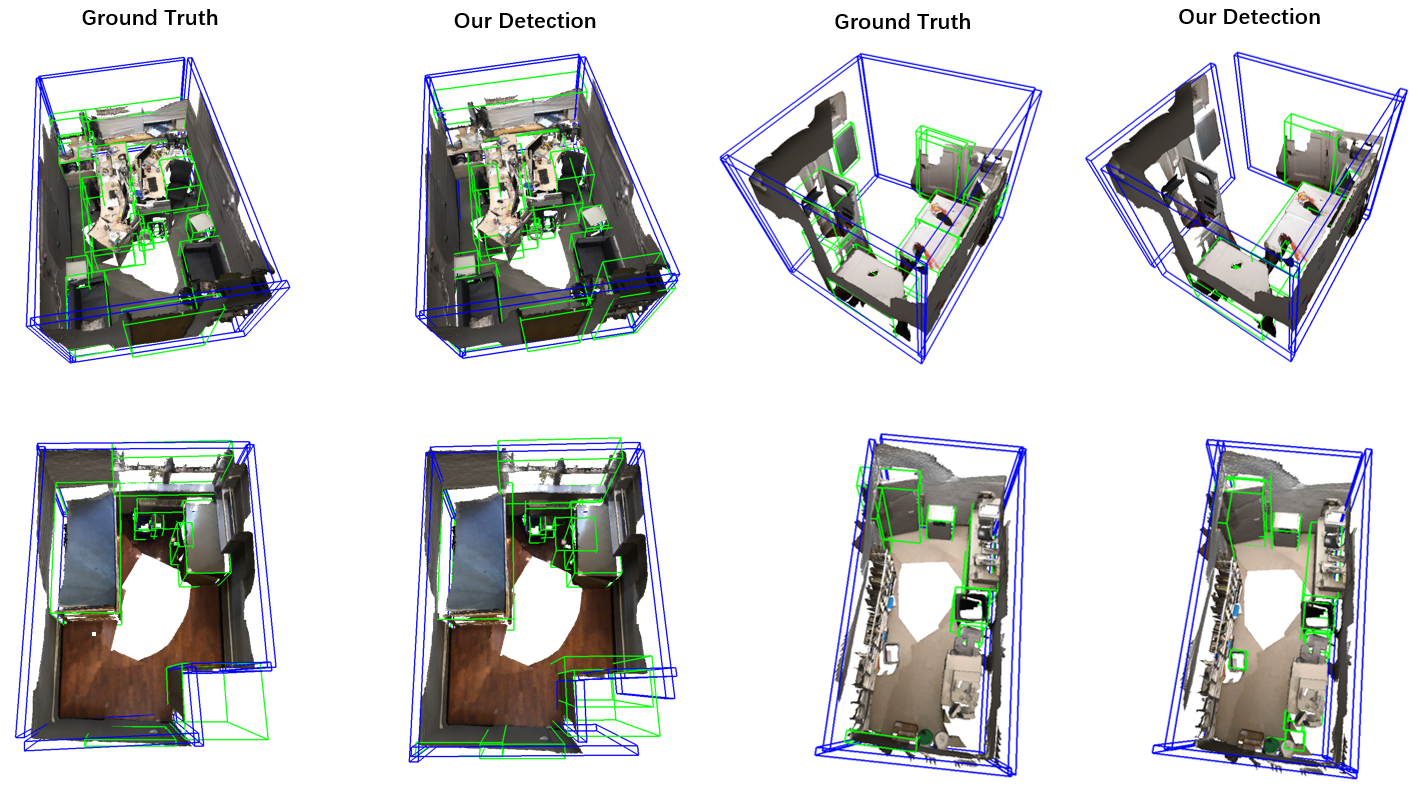}}
  \subfigure{
    \label{fig:subfig:subfig-b} 
    \includegraphics[scale=0.5]{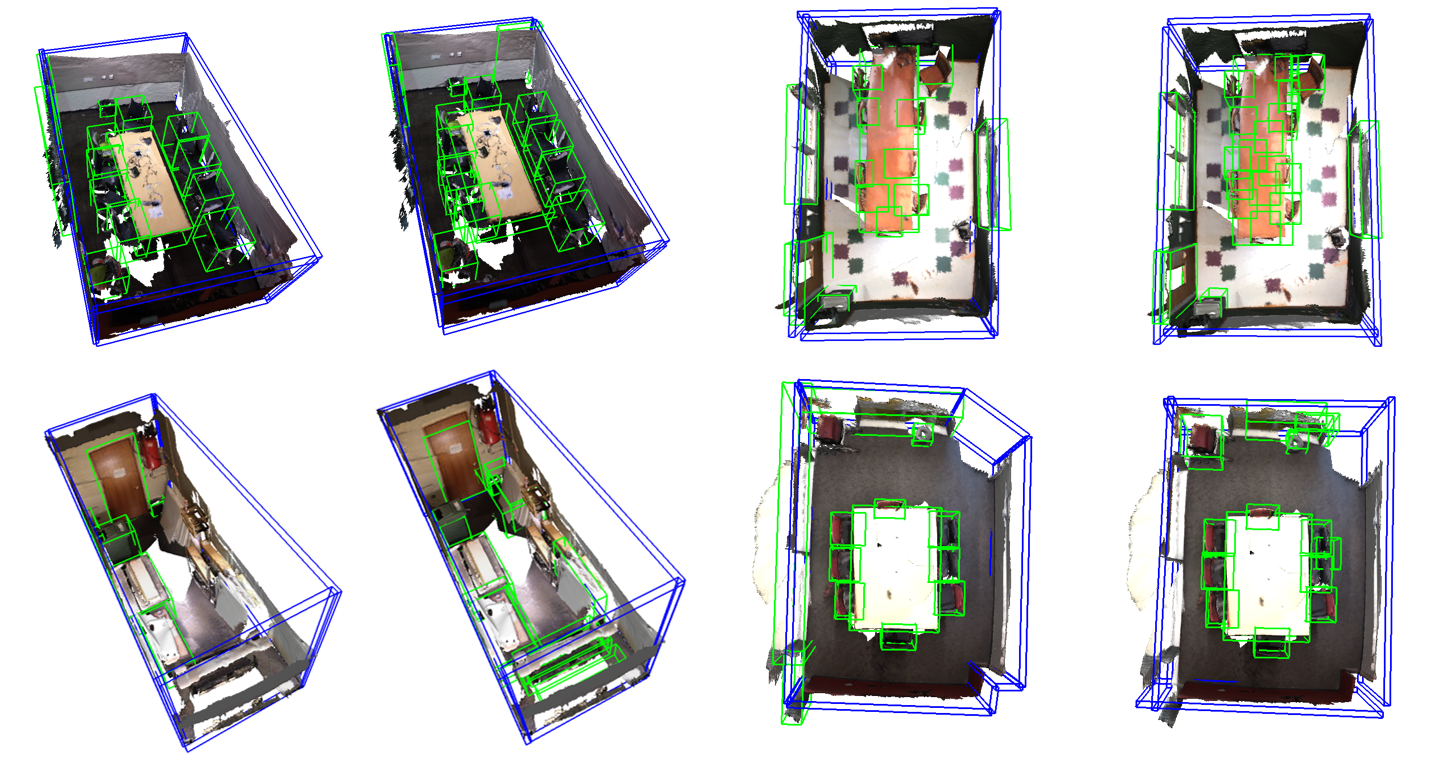}}
  \subfigure{
    \label{fig:subfig:subfig-b} 
    \includegraphics[scale=0.5]{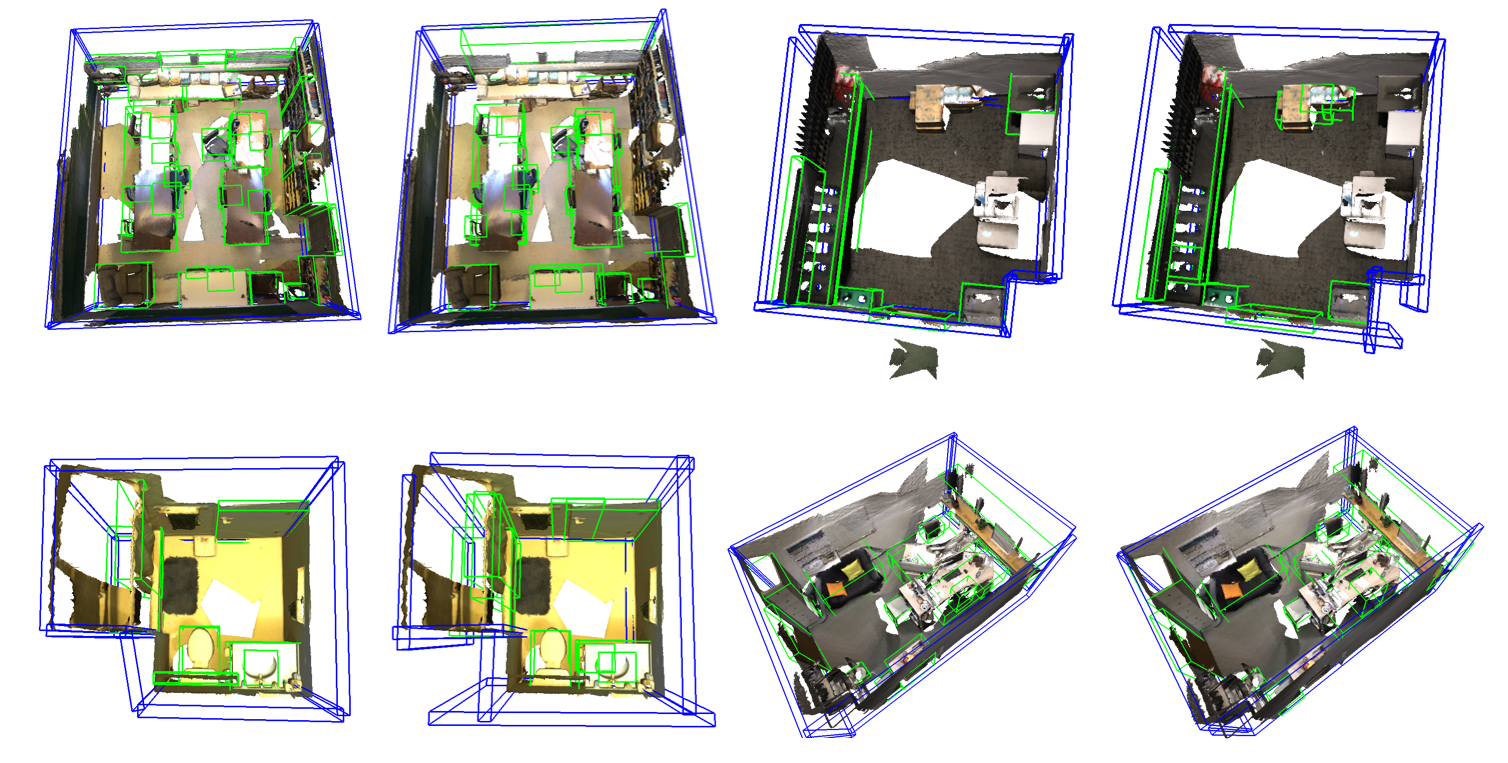}}
    \caption{More qualitative results.}
\end{figure*}

\begin{figure*}[t]
  \centering
  \ContinuedFloat
    \subfigure{
    \label{fig:subfig:subfig-a} 
    \includegraphics[scale=0.5]{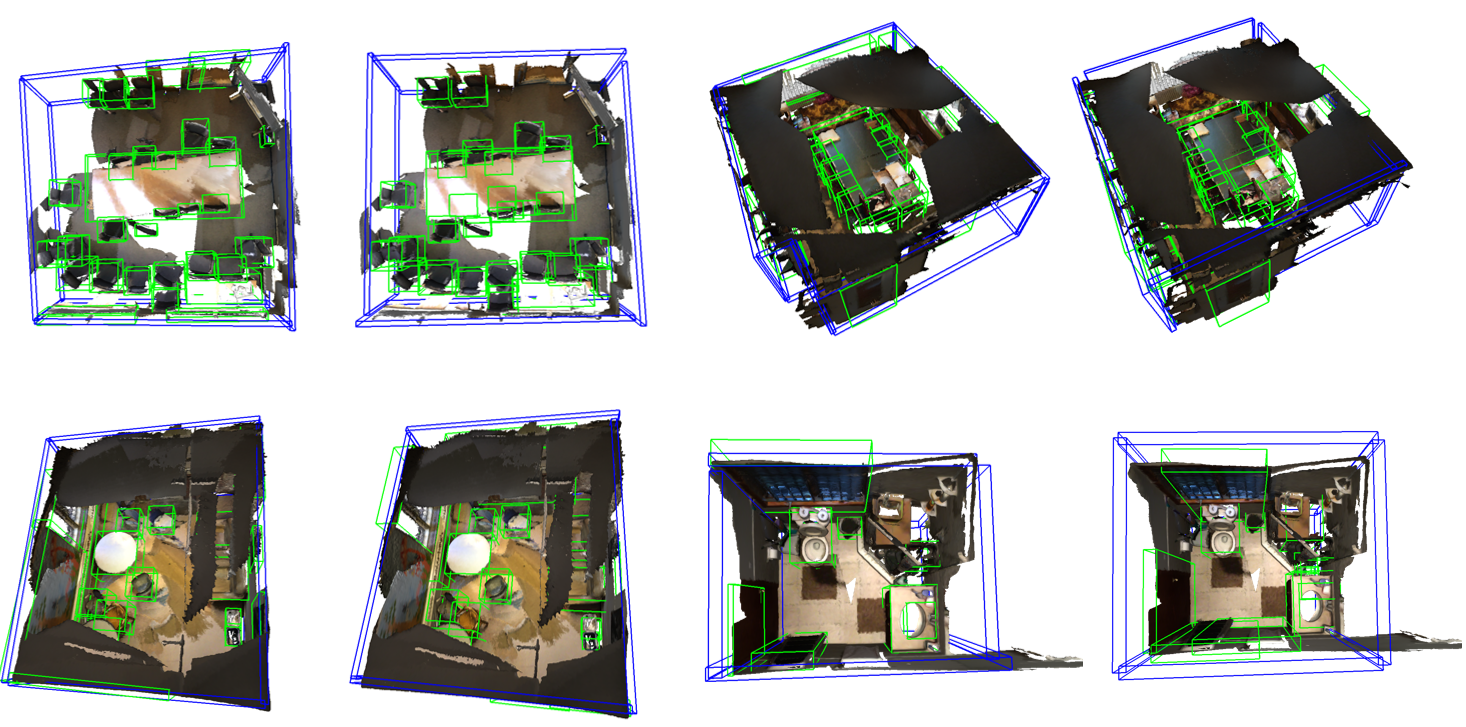}}
  \subfigure{
    \label{fig:subfig:subfig-b} 
    \includegraphics[scale=0.5]{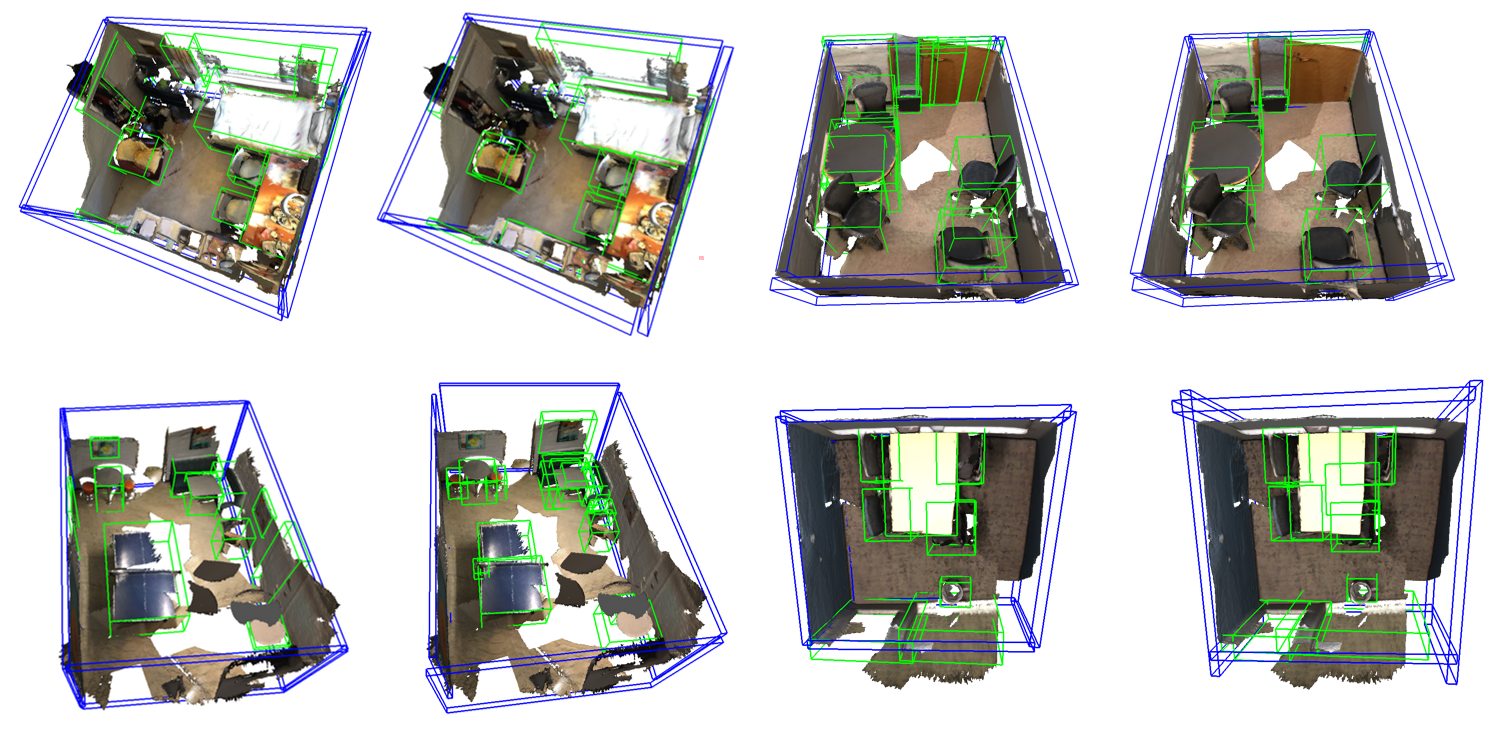}}
  \caption{More qualitative results (cont.).}
  \label{fig:more} 
\end{figure*}

\begin{table*}[b]
\caption{Per-category 3D object detection results on ScanNet.}
\begin{center}
\begin{tabular}{ccccccccccc}
\hline
\multicolumn{1}{c|}{Method}& bathtub & bed & bshelf & cabinet& chair & counter & curtain & desk & door & \multicolumn{1}{|c}{mAP@0.25} \\
\cline{1-11} 
\multicolumn{1}{c|}{VoteNet \cite{b1}}  & 92.1 & 87.9 & 44.3 & 36.3 & 88.7 & 56.1  & 47.2 & 71.7 & 47.3 & \multicolumn{1}{|c}{58.7}
 \\
\multicolumn{1}{c|}{H3DNet \cite{b3}}  & \textbf{92.5} & 88.6 & 54.9 & 49.4 & 91.8 & \textbf{62.0}  & 57.3 & 75.9  & 55.8 & \multicolumn{1}{|c}{67.2}
 \\
\multicolumn{1}{c|}{Group-Free (L6,O256) \cite{b4}}  & \textbf{92.5} & 86.2 & 48.5 & 54.1 & \textbf{92.0} & 59.4  & 64.2 & 80.4 & 55.8 & \multicolumn{1}{|c}{\textbf{67.3}}    \\
\cline{1-11} 
\multicolumn{1}{c|}{Ours (joint, one proposal)}  & 50.5 &  79.3 & 28.3 & 35.7 &  75.8 & 17.5 & 41.2 & 60.0 & 27.8 & \multicolumn{1}{|c}{44.6}   \\
\multicolumn{1}{c|}{Ours (joint, w/o $\mathcal{L}_{pc}$)}  & 90.9 & 89.6 & 43.0 & 42.6 & 87.4 &  61.4 & \textbf{69.3} & 77.5 & 51.7 & \multicolumn{1}{|c}{64.4}   \\
\multicolumn{1}{c|}{Ours (joint)}  & 90.0 & \textbf{94.4} & \textbf{65.3} & \textbf{55.2} & 89.5 & 51.3  & 58.7 & \textbf{87.5} & 58.4& \multicolumn{1}{|c}{66.9}   \\
\multicolumn{1}{c|}{Ours (single)}& 88.5 & \textbf{94.4} & 54.2 & 50.0 & 88.2 & 55.3 & 64.6  & 84.3 & \textbf{60.5} & \multicolumn{1}{|c}{67.2} \\
\cline{1-11} 
\multicolumn{1}{c|}{Method}& gbin & picture &fridge& sink & scurtain & sofa & table & toilet & window \\
\cline{1-10}
\multicolumn{1}{c|}{VoteNet \cite{b1}}  & 37.2 & 7.8 & 45.4 & 54.7 & 57.1 & 89.6  & 58.8 & 94.9  & 38.1  \\
\multicolumn{1}{c|}{H3DNet \cite{b3}}  & 53.6 & 18.6 & 57.2 & 67.4 & 75.3 & 90.2  & 64.9 & 97.9  & \textbf{51.9}  \\
\multicolumn{1}{c|}{Group-Free (L6,O256) \cite{b4}}  & \textbf{55.0} & 15.0 & 57.2 & \textbf{76.8} & 76.3 & 84.8  & \textbf{67.8} & 97.6 & 46.9  \\
\cline{1-10} 
\multicolumn{1}{c|}{Ours (joint, one proposal)} & 26.1 & 3.5 & 28.5 & 65.3 & 48.8 & 76.8 & 20.6 & 89.0 & 28.6 \\
\multicolumn{1}{c|}{Ours (joint, w/o $\mathcal{L}_{pc}$)}  & 45.8 & 15.5 & 56.0 & 64.5 & 79.0 &  96.6 & 47.0 & 96.3 & 44.9 \\
\multicolumn{1}{c|}{Ours (joint)}  & 53.5 & 14.9 & \textbf{60.0} & 62.1 & 65.4& \textbf{96.9} & 54.7& 97.6 & 48.1   \\
\multicolumn{1}{c|}{Ours (single)} & 54.1 & \textbf{21.8} & 54.2 & 65.8 & \textbf{81.1} & 90.0 & 51.6  & \textbf{98.4} & \textbf{51.9} \\
\cline{1-10}
\end{tabular}
\label{objectdetection}
\end{center}
\end{table*}


\end{document}